\documentclass[10pt,twocolumn,letterpaper]{article}

\usepackage{cvpr}              
\definecolor{cvprblue}{rgb}{0.21,0.49,0.74}
\usepackage[pagebackref,breaklinks,colorlinks,allcolors=cvprblue]{hyperref}
\usepackage{booktabs}  
\usepackage{pifont}    
\usepackage{caption}   
\usepackage{array}     
\usepackage{multirow}  
\usepackage{cuted}     
\usepackage{colortbl}
\usepackage{xcolor}
\definecolor{cvprblue}{RGB}{0, 82, 147}
\usepackage{arydshln}
\usepackage{graphicx}
\usepackage{lipsum}
\usepackage{indentfirst}
\usepackage{amsmath}
\usepackage{cases}
\usepackage{booktabs}
\usepackage{multirow} 
\usepackage{graphicx}
\usepackage{times}  
\usepackage{algorithm}
\usepackage{algorithmic}
\usepackage{color}
\usepackage{ulem}
\usepackage{pifont}
\usepackage{arydshln} 
\usepackage{float} 
\usepackage{bm}
\usepackage{caption}
\usepackage{listings}
\def\paperID{*****} 
\def\confName{CVPR}
\def\confYear{2026}

\title{UniSwap: Streaming Audio-Visual Identity Swapping for Talking Videos}

\author{
Yuxuan Zhang \textsuperscript{1,2}
\quad Haozhong Xiong \textsuperscript{2} 
\quad Jiayi Song \textsuperscript{2}
\quad Jinpeng Yu \textsuperscript{2} \\
\quad Yang Shi \textsuperscript{2}
\quad Jiaming Liu \textsuperscript{2} 
\quad Ruihua Huang \textsuperscript{2} 
\quad Liwei Wang \textsuperscript{1}
\\
\textsuperscript{1} The Chinese University of Hong Kong \\
\textsuperscript{2} Qwen Applications Business Group of Alibaba\\
{\tt\small yxzhang@cse.cuhk.edu.hk, jmliu1217@gmail.com} \\
{\url{https://uniswap-av.github.io/}}
}

\begin{document}
\maketitle

\begin{strip}
    \centering
    \includegraphics[width=0.9\textwidth]{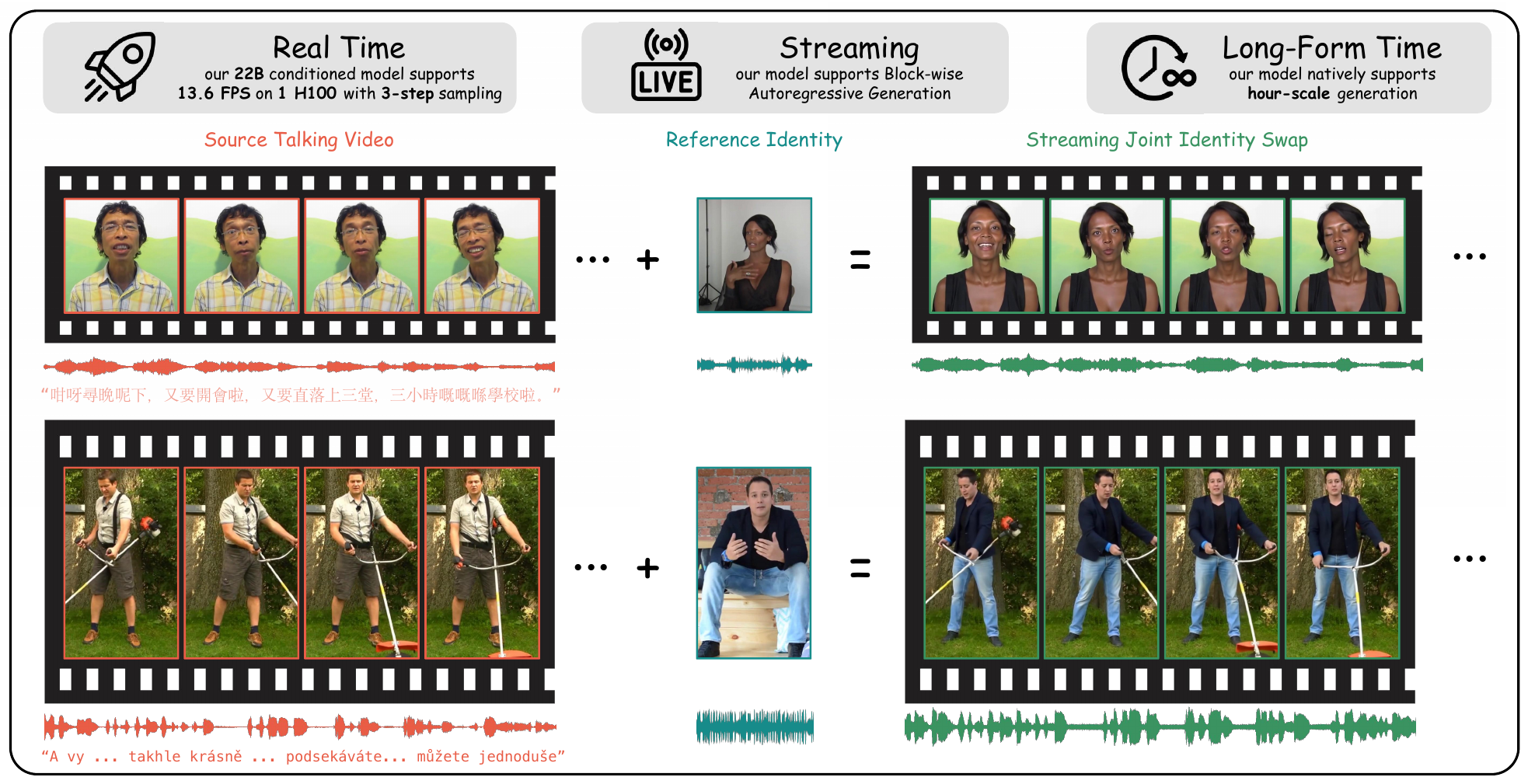}
    \captionof{figure}{\textbf{UniSwap performs streaming joint audio-visual identity replacement.} Given a source talking video and a reference identity (an image and voice clip), UniSwap jointly transfers appearance and vocal timbre while preserving the source motion, background, and linguistic content. Blockwise autoregressive generation with 3-step sampling runs at 13.6 FPS on one NVIDIA H100 and supports stable hour-scale long-form generation.}
    \label{fig:teaser}
\end{strip}

\begin{abstract}

Talking-video character replacement requires coordinated transfer of appearance and voice while preserving the source motion, scene, linguistic content, and audio-video timing. Existing methods use separately optimized models for the two modalities, making audio-visual consistency difficult to enforce. We present \textbf{UniSwap}, the first framework for streaming joint audio-visual identity replacement in talking videos. Given a source video, a reference image, and a reference voice clip, UniSwap transfers the reference appearance and vocal timbre within a single audio-visual diffusion transformer while preserving the source content and dynamics. To address the scarcity of aligned cross-identity training pairs, we introduce a swap-and-reconstruct pipeline that removes visual and vocal identity from real clips and uses the original clips as reconstruction targets. Starting from a bidirectional backbone, we progressively adapt the model through In-context Pretraining for joint replacement, Conditional Streaming Adaptation for block-causal KV-cached generation, and Efficient Self-forcing DMD for mitigating exposure bias and reducing sampling from 30 to 3 denoising steps per block. Efficient Multi-LoRA Switching enables the three DMD roles to share a single frozen backbone. Feature-RoPE Decomposition keeps cached positions within the training range, supporting stable long-form inference. Experiments demonstrate strong audio-visual synchronization, competitive identity preservation, efficient streaming, and stable long-form generation.

\end{abstract}

\section{Introduction}
\label{sec:intro}

Recent generative models have substantially advanced video synthesis~\cite{wan2025,ltxvideo,cogvideox} as well as speech generation and neural audio modeling~\cite{cosyvoice,encodec}. These advances enable applications such as film post-production, content localization, and personalized media. A central capability in these applications is audio-visual identity replacement: changing both the appearance and voice of a person in a talking video while preserving the source motion, background, and linguistic content. Performing this transformation jointly within a single streaming model would enable low-latency, interactive workflows. However, existing methods are not designed to jointly replace visual and vocal identity while supporting streaming generation.

Existing methods address only one side of this task. Video character-replacement methods~\cite{mocha,wananimate,vace,hunyuancustom} transfer appearance from a reference image but do not generate the corresponding voice. Conversely, voice-conversion systems~\cite{seedvc,refvc,cosyvoice} modify speaker timbre without visual context. Combining these components in a cascade can replace both identities, but the two modules remain independently optimized: no joint objective enforces consistency between the converted speech and lip motion, and errors from one modality cannot be corrected using evidence from the other. In addition, most diffusion-based video replacement methods require the complete clip before generation, preventing blockwise streaming and limiting their use in low-latency interactive applications.

We therefore formulate character replacement in talking videos as joint audio-video generation. A unified model can condition on both target appearance and vocal timbre while using cross-modal interactions to coordinate generated speech and lip motion. Realizing this formulation presents two challenges. First, training requires source--target pairs that differ in visual and vocal identity while remaining aligned in motion, scene content, speech, and timing; such pairs are impractical to collect at scale. Second, high-quality audio-video diffusion backbones operate bidirectionally over complete sequences and require many denoising steps, whereas low-latency streaming requires causal, blockwise generation with bounded per-block computation.

To address these challenges, we introduce \textbf{UniSwap}, the first framework for \textit{streaming joint audio-visual identity replacement in talking videos} (Fig.~\ref{fig:teaser}). Given a source talking video, a reference image, and a reference voice clip, UniSwap generates synchronized video and audio in which the reference appearance and vocal timbre are transferred while the source motion, background, and linguistic content are preserved. Unlike cascaded systems, UniSwap performs both replacements within a single audio-video diffusion transformer and generates the output autoregressively in blocks.

To construct aligned supervision, our swap-and-reconstruct pipeline retains each real clip as the target while synthesizing an identity-altered source: an identity-reduced motion proxy replaces the person, and voice conversion changes the speaker timbre without altering the speech content or timing. The original appearance and voice provide the references. This process converts ordinary talking videos into aligned training pairs without requiring different people to perform the same motion and speech.

UniSwap builds on LTX-2.3~\cite{ltxvideo}, an audio-video diffusion transformer with native cross-modal attention, and progressively adapts it through three training stages. Stage 1, In-context Pretraining, concatenates the source, reference, and noisy target latents into a unified sequence with aligned source--target coordinates, allowing full-sequence attention to learn joint visual and vocal identity replacement. Stage 2, Conditional Streaming Adaptation, applies a Decoupled Streaming Conditioning Mask that restricts each token region to its inference-time receptive field, converting the bidirectional model into a block-causal generator and enabling KV-cached autoregressive inference. Stage 3, Efficient Self-forcing DMD, rolls out the student's own predictions to expose it to inference-like histories, while DMD~\cite{dmd} reduces sampling from 30 to 3 denoising steps per block. Efficient Multi-LoRA Switching lets the teacher, generator, and critic share one frozen backbone instead of maintaining three full model copies. During inference, Feature-RoPE Decomposition separates cached features from rotary coordinates and reapplies bounded positions, preserving cross-modal temporal alignment and supporting stable long-form generation.


Our main contributions are fourfold:
\begin{itemize}
    \item We introduce UniSwap, the first framework for streaming joint audio-visual identity replacement in talking videos, transferring appearance and vocal timbre within a single diffusion transformer.
    \item We propose a swap-and-reconstruct pipeline that converts ordinary talking videos into temporally aligned supervision for joint visual and vocal identity replacement.
    \item We progressively convert a bidirectional backbone into a causal three-step generator through In-context Pretraining, Conditional Streaming Adaptation with a Decoupled Streaming Conditioning Mask, and Efficient Self-forcing DMD implemented through Efficient Multi-LoRA Switching.
    \item We propose Feature-RoPE Decomposition, which separates cached features from rotary coordinates and bounds positional indices while preserving cross-modal temporal alignment during long-form inference.
\end{itemize}

\section{Related Work}
\label{sec:related}

\subsection{Video Character Replacement}

Video character replacement substitutes a person's visual identity while
preserving motion, expression, and background. Frame-based face-swapping
approaches~\cite{simswap,faceshifter,infoswap} can suffer from temporal
inconsistency. Recent methods instead use video diffusion models.
MoCha~\cite{mocha} casts replacement as in-context generation and uses
condition-aware RoPE~\cite{rope} to combine a source video, a mask, and reference
images without per-frame structural guidance. Wan-Animate~\cite{wananimate} and
VACE~\cite{vace} inject explicit structural signals into
DiT-based~\cite{dit} generation, while HunyuanCustom~\cite{hunyuancustom}
conditions customized generation on masked target regions and reference images.
More generally, in-context diffusion methods place conditioning examples and
noisy targets in a shared transformer context. IC-LoRA~\cite{iclora} studies
this formulation for image diffusion transformers, and Video Diffusion
Transformers are In-Context Learners~\cite{videoincontext} extends it to video
generation by concatenating related videos along spatial or temporal
dimensions. FullDiT2~\cite{fulldit2} further improves the efficiency of
in-context video conditioning through dynamic token selection and selective
context caching. These methods address visual identity or visual control only,
so dubbing still requires a separate audio system.

\subsection{Voice Conversion}

Voice conversion changes a speaker's timbre while preserving linguistic content
and prosody. Traditional approaches rely on parallel data or explicit feature
disentanglement, whereas recent zero-shot
methods~\cite{seedvc,refvc,cosyvoice} use large-scale pretraining for
speaker-independent conversion. Seed-VC~\cite{seedvc} uses self-supervised
speech representations for content extraction. REF-VC~\cite{refvc} combines
ASR bottleneck and self-supervised features in a diffusion transformer, with
random erasing and shortcut models for robust, fast inference.
CosyVoice~\cite{cosyvoice} uses supervised semantic tokens for zero-shot
text-to-speech synthesis. These systems operate without visual context and
therefore do not jointly optimize converted speech and lip motion.

\subsection{Audio-Visual Generation}

Recent generative models have begun to model audio and video jointly. An early
representative, MM-Diffusion~\cite{mmdiffusion}, couples modality-specific
denoisers through cross-modal attention to synthesize aligned audio-video pairs.
More recent diffusion transformers strengthen joint modeling:
JavisDiT~\cite{javisdit} introduces hierarchical spatio-temporal priors for
audio-visual synchronization, while LTX-2~\cite{ltxvideo} uses separate,
interacting modality streams connected by cross-modal attention, enabling
synchronized audio-video generation from text. UniSwap builds on this capability
for character replacement, where the model must transfer visual and vocal
identity while retaining source motion and speech content.

\subsection{Distillation for Real-Time Generation}

Diffusion models typically require many denoising steps, limiting low-latency applications. Distribution Matching Distillation (DMD)~\cite{dmd,dmd2} trains a student to match a multi-step teacher's output distribution using a critic, enabling one- or few-step generation. Consistency models~\cite{consistencymodel} instead enforce consistency along the probability-flow ODE trajectory. For continuous sequence generation, Diffusion Forcing~\cite{diffusionforcing} combines causal prediction with diffusion over independently noised sequence tokens. For autoregressive video generation, Self-Forcing~\cite{selfforcing} reduces exposure bias by feeding the model's own outputs back during training, while CausVid~\cite{causvid} applies DMD to causal streaming video. Rolling Forcing~\cite{rollingforcing} combines rolling-window
denoising, an initial-frame attention sink, and few-step distillation for long-horizon video streams. OmniForcing~\cite{omniforcing} further distills a
bidirectional dual-stream audio-visual diffusion model into a streaming autoregressive generator using block-causal alignment, joint self-forcing distillation, and a rolling KV cache. Unlike these general generation systems, UniSwap combines self-forcing rollout and DMD for source-driven joint appearance-and-voice streaming replacement.


\begin{figure*}[!ht]
    \centering
    \includegraphics[width=\textwidth]{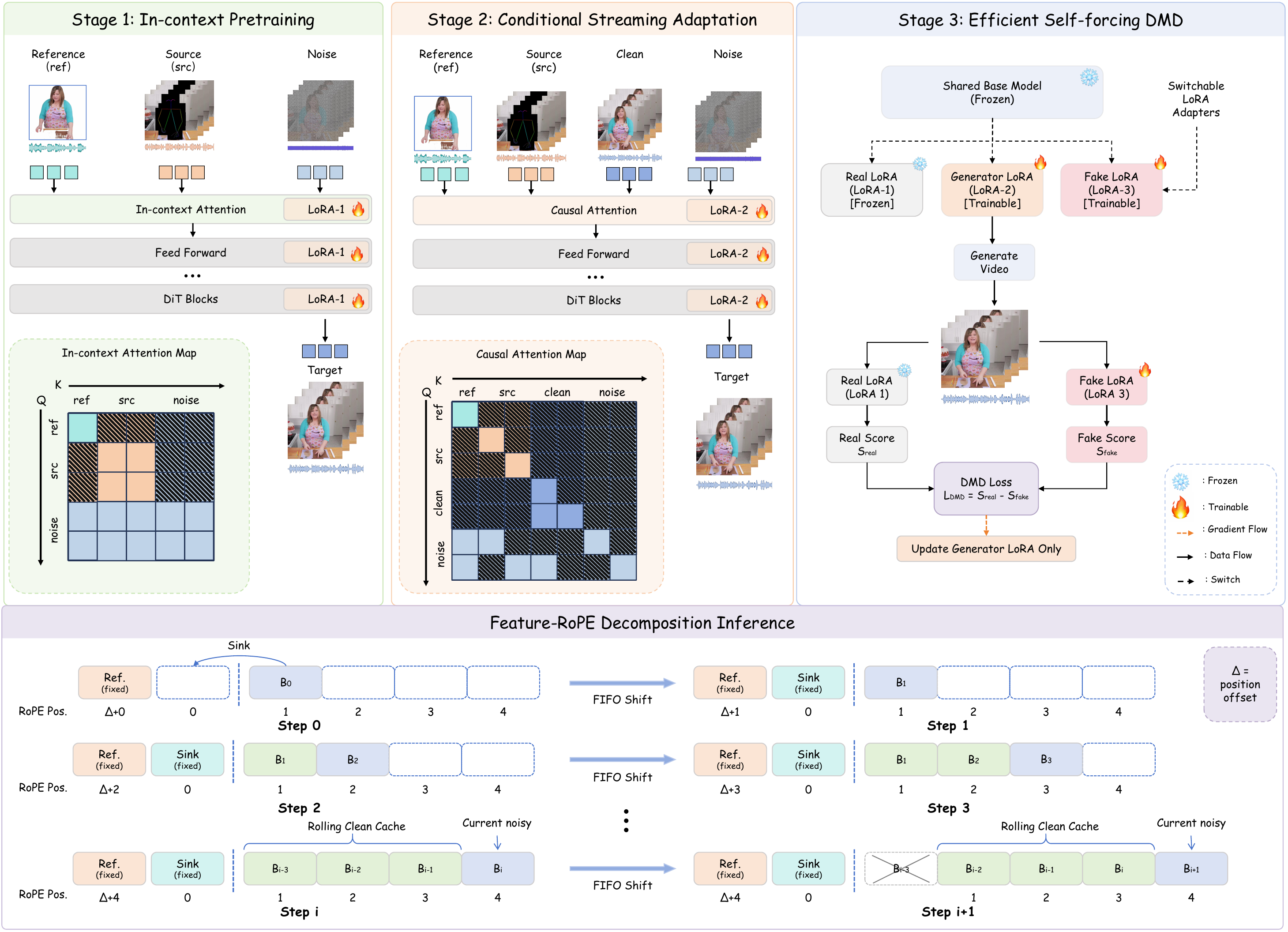}
    \caption{\textbf{Overview of UniSwap.} The three-stage pipeline comprises
    (a) In-Context Pretraining for joint audio-video replacement,
    (b) Conditional Streaming Adaptation with block-causal masking and
    KV-cached inference, and (c) Efficient Self-Forcing DMD, which reduces
    denoising to 3 steps per block. Feature-RoPE Decomposition bounds cached
    positions while preserving cross-modal physical-time alignment.}
    \label{fig:pipeline}
\end{figure*}

\section{Method}
\label{sec:method}

Given a source video $V_s \in \mathbb{R}^{F \times 3 \times H \times W}$ with audio waveform $A_s$, a reference image $I_r$, and a reference voice clip $A_r$, UniSwap generates a target video $V_t$ and audio $A_t$. The target should match the reference appearance and vocal timbre while preserving the source motion, lip movements, background, and linguistic content. Figure~\ref{fig:pipeline} illustrates the three-stage training pipeline.

\noindent\textbf{Backbone and Latent Representation.}
We build on the frozen LTX-2.3 audio-video diffusion transformer~\cite{ltxvideo}, which processes video and audio in separate streams connected by bidirectional cross-modal attention. A causal video VAE compresses the video temporally by a factor of eight, while the audio representation is sampled at 25 latent tokens per second. We denote the video and audio latent channel dimensions by $C_v$ and $C_a$, respectively; their token positions are assigned on a shared physical-time axis to maintain cross-modal alignment.

\subsection{Swap-and-Reconstruct Paired Data Synthesis}
\label{sec:data}

\begin{figure*}[!t]
    \centering
    \includegraphics[width=\textwidth]{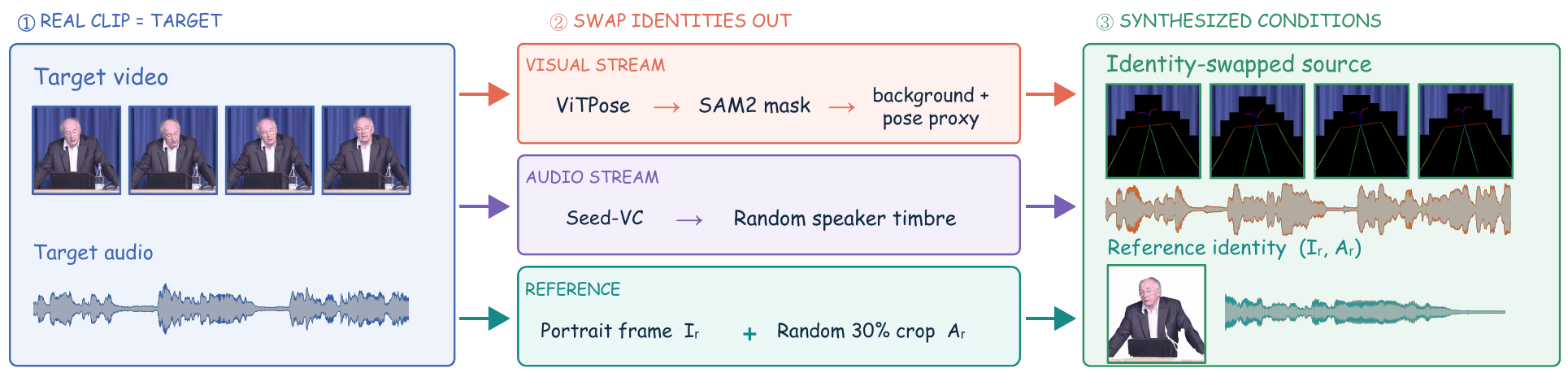}
    \caption{\textbf{Swap-and-reconstruct paired data synthesis.} Every real talking video serves as its own reconstruction target. \ding{172} The real clip provides the target video $V_t$ and audio $A_t$. \ding{173} The visual identity is swapped out by replacing the person with a pose proxy composited onto the masked background plate, the vocal timbre is randomized towards a sampled speaker, and the reference identity is formed by a portrait frame $I_r$ and a random $30\%$ audio crop $A_r$. \ding{174} The resulting identity-swapped source and reference identity condition the model, which is trained to reconstruct the original clip.}
    \label{fig:data_pipeline}
\end{figure*}

Paired data for joint audio-video identity replacement are not naturally available at scale. We therefore formulate the task as self-reconstruction: a real talking video provides the target $(V_t,A_t)$, while an identity-altered version provides the source $(V_s,A_s)$. The original visual and vocal identities are supplied separately through a reference image $I_r$ and reference audio $A_r$ (Fig.~\ref{fig:data_pipeline}).

\noindent\textbf{Visual Stream.}
We estimate whole-body 2D poses~\cite{vitpose} and use the pose keypoints to prompt video segmentation~\cite{sam2}. After dilating and augmenting the person mask, we remove the person to obtain a background plate and composite the rendered pose sequence onto it, yielding $V_s$. This retains the source motion, timing, and background while suppressing appearance cues; a portrait frame from $V_t$ serves as $I_r$.

\noindent\textbf{Audio Stream.}
We convert $A_t$ toward a randomly sampled speaker using an off-the-shelf voice conversion model~\cite{seedvc}, producing $A_s$ with altered timbre but preserved speech content and prosody. A random segment spanning $30\%$ of $A_t$ serves as the reference audio $A_r$.

All three training stages use tuples generated by this pipeline.

\subsection{Stage 1: In-context Pretraining}
\label{sec:stage1}

Following the in-context conditioning paradigm for diffusion transformers~\cite{videoincontext,iclora}, the first stage adapts the model to joint audio-video character replacement by placing the conditioning and target latents in a unified attention context. As shown in Stage~1 of Fig.~\ref{fig:pipeline}, all inputs are concatenated into a unified token sequence.

\noindent\textbf{Input Formulation.}
For each training sample, we encode: (1) the source video latent $z_s^v \in \mathbb{R}^{L_s^v \times C_v}$ and source audio latent $z_s^a \in \mathbb{R}^{L_s^a \times C_a}$ representing the driving motion and content; (2) a reference image latent $z_r^v \in \mathbb{R}^{L_r^v \times C_v}$ and reference audio latent $z_r^a \in \mathbb{R}^{L_r^a \times C_a}$ providing the target identity; and (3) the target video latent $z_t^v$ and target audio latent $z_t^a$ to be denoised.

These are concatenated along the sequence dimension for each modality stream:
\begin{equation}
    \text{Video:} \quad x^v = [z_r^v; z_s^v; z_t^v], \quad
    \text{Audio:} \quad x^a = [z_r^a; z_s^a; z_t^a]
\end{equation}
During training, noise is added only to the target portions $z_t^v$ and $z_t^a$; the source and reference tokens serve as clean conditioning context with $\sigma=0$.

\noindent\textbf{Condition Positional Encoding Offset.}
A naive application of 3D RoPE~\cite{rope} would assign sequential temporal indices to all tokens, coupling the model to fixed input lengths. Instead, we employ condition positional encoding offsets: the source and target share identical temporal positions in both modalities because they represent the same time span, while the reference image and reference audio use fixed offsets $\Delta_r^v$ and $\Delta_r^a$, respectively. This design enables variable-length inputs while keeping the positional semantics of all conditions consistent throughout adaptation and inference.

\noindent\textbf{Training.}
The model is trained end-to-end with in-context attention across the entire concatenated sequence. For a clean latent $z_0$, Gaussian noise $\epsilon$, and noise level $\sigma$, we define $z_\sigma=(1-\sigma)z_0+\sigma\epsilon$ and the conditional flow-matching loss as
\begin{equation}
    \mathcal{L}_{\mathrm{FM}} =
    \mathbb{E}_{z_0,\sigma,\epsilon}
    \left[\left\|v_\theta(z_\sigma,\sigma)-(\epsilon-z_0)\right\|_2^2\right].
\end{equation}
Both video and audio target streams are jointly denoised:
\begin{equation}
    \mathcal{L}_{\text{Stage1}} = \mathcal{L}_{\text{FM}}^{\text{video}} + \mathcal{L}_{\text{FM}}^{\text{audio}}
\end{equation}

\subsection{Stage 2: Conditional Streaming Adaptation}
\label{sec:stage2}

Stage 2 converts the bidirectional model into a blockwise autoregressive generator. We partition the target into blocks of $K{=}3$ video latent frames, except for an initial 4-frame block required by the temporal indexing of the causal video VAE. Audio blocks are defined over the corresponding time intervals. Let $B_i$ and $S_i$ denote the $i$-th target and source blocks, respectively. The model denoises $B_i$ conditioned on the reference, $S_i$, and the preceding target blocks $B_{<i}$.

\subsubsection{Decoupled Streaming Conditioning Mask}
The Decoupled Streaming Conditioning Mask aligns the training-time receptive field with KV-cached inference (Stage~2 of Fig.~\ref{fig:pipeline}). Reference tokens and each source block are encoded independently. Clean target blocks follow block-causal attention, whereas the noisy target block $B_i$ attends to the reference, its aligned source block $S_i$, the clean history $B_{<i}$, and itself. The same role-wise mask is applied to audio and video self-attention and to cross-modal attention, with modality-specific block sizes chosen to preserve physical-time alignment. This construction prevents future-target leakage while reproducing the context available during autoregressive inference.


During training, the clean stream is populated with ground-truth latents (teacher forcing~\cite{teacherforcing}), so the model learns to denoise each block $B_i$ under ground-truth history. The loss is computed only on the noisy target tokens:
\begin{equation}
    \mathcal{L}_{\text{Stage2}} = \mathbb{E}_{i, \sigma} \left[ \mathcal{L}_{\text{FM}}(B_i \mid \text{ref}, S_i, B_0^{\text{clean}}, \ldots, B_{i-1}^{\text{clean}}) \right]
\end{equation}

\noindent\textbf{KV-Cached Streaming Inference.}
At inference, the reference is cached once, each source block is cached only while its target block is denoised, and completed target blocks are committed as clean history. This yields a per-block cost independent of the generated duration. The complete cache update procedure is provided in Alg.~\ref{alg:kv_cache} of the supplementary material.

\newcommand{\experimentdisplays}{%
\begin{figure*}[!ht]
    \centering
    \includegraphics[width=\textwidth]{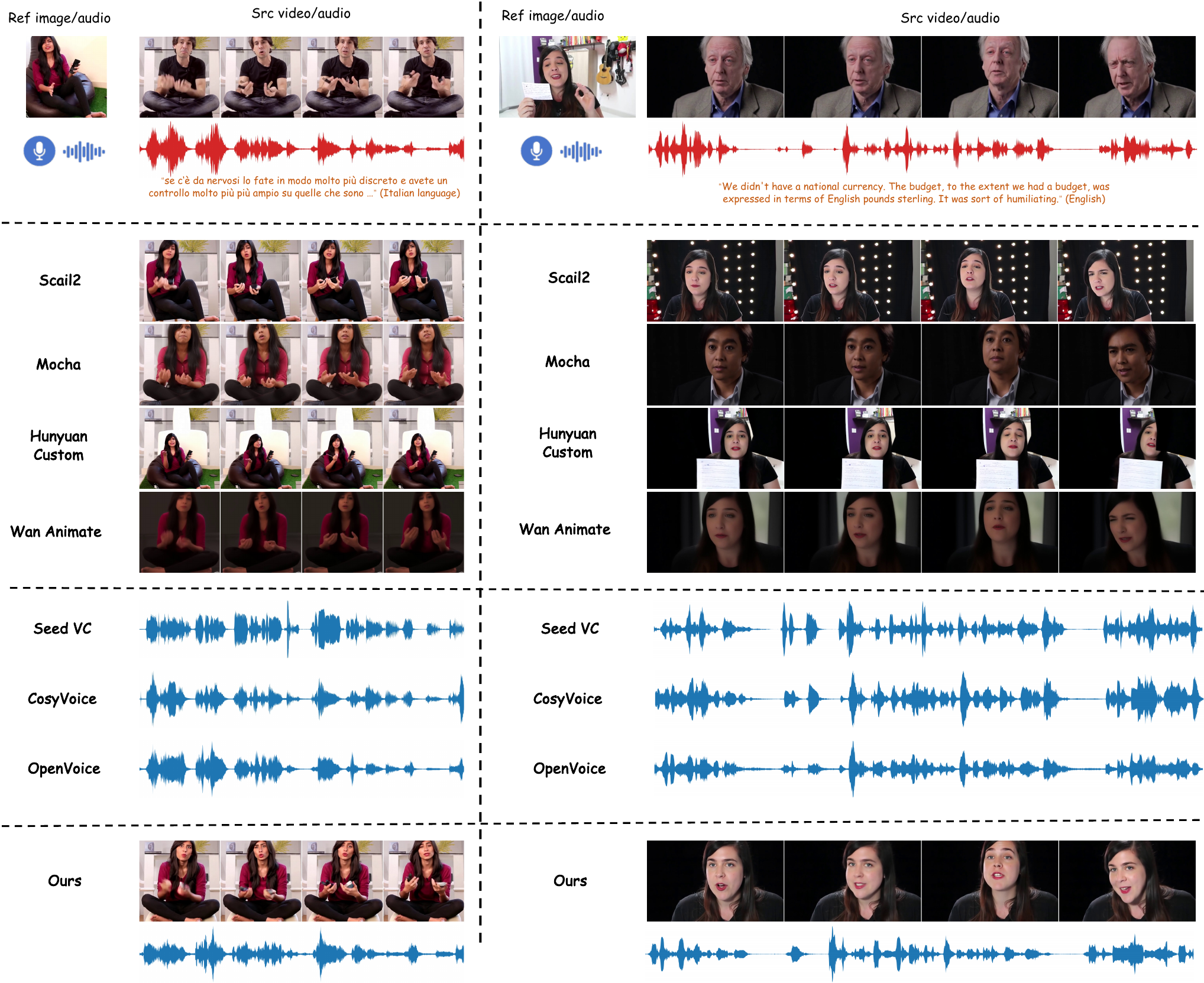}
    \caption{\textbf{Qualitative comparison on the short-video benchmark.} Given the reference image/audio and the source video/audio (top), video replacement methods (upper rows) transfer the appearance but keep the original voice, and voice conversion methods (middle rows) modify only the speech waveform, while UniSwap (bottom) jointly replaces the appearance and the voice, producing lip motion synchronized with the converted speech. Zoom in for details.}
    \label{fig:qual_short}
\end{figure*}

\begin{figure*}[!ht]
    \centering
    \includegraphics[width=\textwidth]{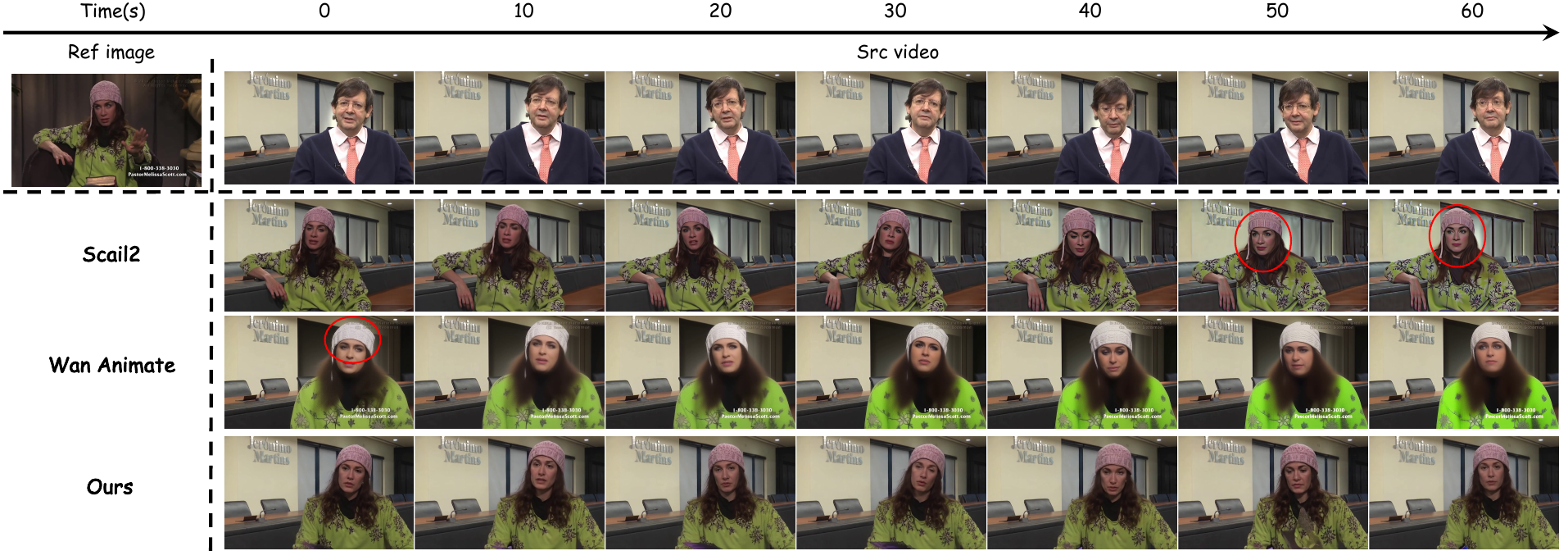}
    \caption{\textbf{Qualitative comparison on the long-video benchmark.} Frames are sampled every 10 seconds from 1-minute generations. The baselines exhibit identity drift and visual artifacts as generation progresses (highlighted in red), while UniSwap preserves the reference identity throughout the full duration. Zoom in for details.}
    \label{fig:qual_long}
\end{figure*}

\begin{table*}[!ht]
\centering
\caption{\textbf{Quantitative comparison on the short-video benchmark.} Each video replacement method is paired with the same Seed-VC audio backend, which matches the converter used to synthesize our training sources; audio-visual synchronization is measured on the resulting cascade, and video quality on the generated video. Because all video replacement baselines share the same Seed-VC output, their repeated voice-quality values are omitted. Voice-conversion methods retain the source video. The best result in each column is in \textbf{bold}, and the second best is \underline{underlined}. ``--'' denotes a metric that is not separately reported for that row.}
\label{tab:main_results}
\vspace{2mm}
\resizebox{\textwidth}{!}{
\begin{tabular}{ll|cc|ccc|ccccc}
\toprule
\multicolumn{2}{l|}{\multirow{2}{*}{Method}} & \multicolumn{2}{c|}{A--V Sync} & \multicolumn{3}{c|}{Video Quality} & \multicolumn{5}{c}{Voice Quality} \\
\cmidrule(lr){3-4} \cmidrule(lr){5-7} \cmidrule(lr){8-12}
& & Sync-C $\uparrow$ & Sync-D $\downarrow$ & ASE $\uparrow$ & IQA $\uparrow$ & DINO-S $\uparrow$ & SIG $\uparrow$ & BAK $\uparrow$ & OVRL $\uparrow$ & SECS $\uparrow$ & SSIM $\uparrow$ \\
\midrule
\multirow{5}{*}{Video}
& VACE~\cite{vace} & 0.832{\scriptsize$\pm$0.320} & 12.800{\scriptsize$\pm$0.933} & 2.059{\scriptsize$\pm$0.355} & 3.269{\scriptsize$\pm$0.547} & 0.400{\scriptsize$\pm$0.110} & -- & -- & -- & -- & -- \\
& Wan-Animate~\cite{wananimate} & 2.874{\scriptsize$\pm$1.653} & 11.338{\scriptsize$\pm$1.713} & 2.098{\scriptsize$\pm$0.340} & 3.514{\scriptsize$\pm$0.520} & 0.580{\scriptsize$\pm$0.140} & -- & -- & -- & -- & -- \\
& SCAIL-2~\cite{scail2} & \underline{3.289{\scriptsize$\pm$1.592}} & 11.269{\scriptsize$\pm$1.623} & \underline{2.409{\scriptsize$\pm$0.352}} & \underline{4.067{\scriptsize$\pm$0.399}} & \textbf{0.630{\scriptsize$\pm$0.150}} & -- & -- & -- & -- & -- \\
& MoCha~\cite{mocha} & 3.031{\scriptsize$\pm$1.678} & \underline{11.198{\scriptsize$\pm$1.881}} & \textbf{2.534{\scriptsize$\pm$0.340}} & \textbf{4.249{\scriptsize$\pm$0.304}} & 0.577{\scriptsize$\pm$0.147} & -- & -- & -- & -- & -- \\
& HunyuanCustom~\cite{hunyuancustom} & 0.894{\scriptsize$\pm$0.401} & 12.991{\scriptsize$\pm$0.864} & 2.319{\scriptsize$\pm$0.414} & 3.816{\scriptsize$\pm$0.489} & 0.624{\scriptsize$\pm$0.142} & -- & -- & -- & -- & -- \\
\midrule
\multirow{3}{*}{Voice}
& OpenVoice~\cite{openvoice} & -- & -- & -- & -- & -- & 3.458{\scriptsize$\pm$0.415} & 3.438{\scriptsize$\pm$0.663} & 2.910{\scriptsize$\pm$0.486} & 0.755{\scriptsize$\pm$0.066} & \textbf{0.363{\scriptsize$\pm$0.162}} \\
& Seed-VC~\cite{seedvc} & -- & -- & -- & -- & -- & \textbf{3.489{\scriptsize$\pm$0.335}} & \textbf{3.750{\scriptsize$\pm$0.571}} & \textbf{3.074{\scriptsize$\pm$0.445}} & \textbf{0.829{\scriptsize$\pm$0.047}} & 0.212{\scriptsize$\pm$0.115} \\
& CosyVoice~\cite{cosyvoice} & -- & -- & -- & -- & -- & 3.461{\scriptsize$\pm$0.249} & \underline{3.738{\scriptsize$\pm$0.456}} & \underline{3.041{\scriptsize$\pm$0.366}} & \underline{0.802{\scriptsize$\pm$0.051}} & 0.137{\scriptsize$\pm$0.106} \\
\midrule
Joint
& \textbf{UniSwap} & \textbf{3.633{\scriptsize$\pm$1.236}} & \textbf{10.304{\scriptsize$\pm$0.849}} & 2.097{\scriptsize$\pm$0.238} & 3.758{\scriptsize$\pm$0.318} & \underline{0.629{\scriptsize$\pm$0.136}} & \underline{3.486{\scriptsize$\pm$0.327}} & 3.563{\scriptsize$\pm$0.543} & 2.988{\scriptsize$\pm$0.366} & 0.730{\scriptsize$\pm$0.064} & \underline{0.269{\scriptsize$\pm$0.161}} \\
\bottomrule
\end{tabular}
}
\end{table*}
}

\subsection{Stage 3: Efficient Self-forcing DMD}
\label{sec:stage3}

Stage 2 is trained with clean ground-truth histories and requires ${\sim}30$ denoising steps per block. Stage 3 instead distills a 3-step student under self-generated histories, reducing both exposure bias and sampling cost.

\subsubsection{Self-forcing Style Training}

\noindent\textbf{Self-forcing Rollout.} The student autoregressively generates the full audio-video sequence and conditions each block on its preceding outputs. Each block is denoised at three noise levels, $[0.999, 0.757, 0.522]$.

\noindent\textbf{Efficient Multi-LoRA Switching.}
Self-forcing DMD ordinarily requires separate teacher, generator, and critic models. Efficient Multi-LoRA Switching represents these roles with three adapters on a shared frozen backbone (Stage~3 of Fig.~\ref{fig:pipeline}): frozen Stage-1 LoRA-1 serves as the teacher, Stage-2-initialized LoRA-2 as the generator, and randomly initialized LoRA-3 as the trainable critic. Activating one adapter at a time avoids maintaining three backbone copies, reducing peak GPU memory from more than 80~GB (out of memory on an 80-GB GPU) to 65.34~GB under the same configuration. Only LoRA-2 is retained for inference.

\noindent\textbf{Distribution Matching Loss.}
For the generated sequence $\hat{z}$, DMD matches the critic score to the classifier-free-guidance (CFG) enhanced teacher score at a noise level $\sigma \sim \mathcal{U}[0.02,0.98]$:
\begin{equation}
\begin{split}
\nabla_{\hat{z}} & =  \underbrace{D_\phi(\hat{z}_\sigma, \sigma)}_{\text{critic on fake}} \\
& - \underbrace{\left[ T_\psi^{+}(\hat{z}_\sigma,\sigma) + \gamma \left(T_\psi^{+}(\hat{z}_\sigma,\sigma) - T_\psi^{-}(\hat{z}_\sigma,\sigma)\right) \right]}_{\text{CFG-enhanced teacher}}
\end{split}
\end{equation}

where $T_\psi^{+/-}$ are the conditional and unconditional predictions of the bidirectional teacher, $D_\phi$ is the critic prediction, and $\gamma_{\mathrm{video}}=3.0$ and $\gamma_{\mathrm{audio}}=5.0$. This signal updates the generator toward the teacher distribution, while the critic is trained on noised student samples.

\subsubsection{Feature-RoPE Decomposition Inference}
\label{sec:rerope}
Absolute RoPE coordinates grow with an autoregressive stream and eventually exceed the range observed during fixed-length training. Feature-RoPE Decomposition instead stores unrotated keys and reapplies RoPE according to each block's current cache slot, without recomputing its features. The cache contains reference tokens, an initial sink block, and a rolling history of recent clean blocks.

\noindent\textbf{Adaptive Sink Block.}
We adapt the attention-sink mechanism~\cite{kvcache} by retaining the first generated block $B_0$ at a fixed local position. Because $B_0$ establishes the initial appearance and voice, it provides a persistent identity anchor as the rolling history advances.

\noindent\textbf{Reference Re-anchoring.}
Before denoising each block, the stored reference keys are re-rotated relative to the current generation slot, together with the modality-specific offsets $\Delta_r^v$ and $\Delta_r^a$ from Sec.~\ref{sec:stage1}. This keeps the reference at the same relative phase as during training.

\noindent\textbf{Window-Bounded RoPE.}
Rolling blocks are remapped into a fixed window of $W$ local slots. If $\tau(h_i)$ is the physical timestamp of the earliest retained rolling block, a token at time $\tau$ is assigned
\begin{equation}
    \widetilde{p}_i(\tau) = \tau - \tau(h_i),
\end{equation}
while $B_0$ remains fixed and the current block occupies the final slot. When the window is full, the oldest rolling block is evicted and the retained blocks shift locally. We use $W{=}4$: one sink block, two rolling blocks, and one current block. Video and audio coordinates are derived from the same physical-time axis, preserving cross-modal alignment after every shift.

\section{Experiments}
\label{sec:experiments}

\experimentdisplays

\begin{table*}[!ht]
\centering
\caption{\textbf{Long-video comparison.} Metrics are computed independently on three 20-second segments of 1-minute generated videos. The best result in each segment is in \textbf{bold} and the second best is \underline{underlined}. UniSwap maintains stable quality and identity across the full duration, while the baselines fluctuate or degrade over time.}
\label{tab:long_video}
\vspace{2mm}
\resizebox{\textwidth}{!}{
\begin{tabular}{l|ccc|ccc|ccc}
\toprule
\multirow{2}{*}{Method} & \multicolumn{3}{c|}{0--20\,s} & \multicolumn{3}{c|}{20--40\,s} & \multicolumn{3}{c}{40--60\,s} \\
\cmidrule(lr){2-4} \cmidrule(lr){5-7} \cmidrule(lr){8-10}
& ASE $\uparrow$ & IQA $\uparrow$ & DINO-S $\uparrow$ & ASE $\uparrow$ & IQA $\uparrow$ & DINO-S $\uparrow$ & ASE $\uparrow$ & IQA $\uparrow$ & DINO-S $\uparrow$ \\
\midrule
SCAIL-2~\cite{scail2} & \textbf{2.628{\scriptsize$\pm$0.203}} & \textbf{4.426{\scriptsize$\pm$0.265}} & \underline{0.566{\scriptsize$\pm$0.175}} & \textbf{2.765{\scriptsize$\pm$0.197}} & \textbf{4.568{\scriptsize$\pm$0.294}} & \underline{0.538{\scriptsize$\pm$0.154}} & \textbf{2.656{\scriptsize$\pm$0.326}} & \textbf{4.254{\scriptsize$\pm$0.401}} & 0.517{\scriptsize$\pm$0.135} \\
Wan-Animate~\cite{wananimate} & \underline{2.241{\scriptsize$\pm$0.397}} & 3.766{\scriptsize$\pm$0.635} & 0.554{\scriptsize$\pm$0.119} & \underline{2.271{\scriptsize$\pm$0.413}} & 3.741{\scriptsize$\pm$0.701} & 0.533{\scriptsize$\pm$0.115} & 2.238{\scriptsize$\pm$0.364} & 3.628{\scriptsize$\pm$0.714} & \underline{0.528{\scriptsize$\pm$0.114}} \\
\midrule
\textbf{UniSwap} & 2.224{\scriptsize$\pm$0.226} & \underline{3.966{\scriptsize$\pm$0.331}} & \textbf{0.596{\scriptsize$\pm$0.122}} & 2.236{\scriptsize$\pm$0.189} & \underline{4.001{\scriptsize$\pm$0.276}} & \textbf{0.590{\scriptsize$\pm$0.126}} & \underline{2.259{\scriptsize$\pm$0.191}} & \underline{4.032{\scriptsize$\pm$0.254}} & \textbf{0.596{\scriptsize$\pm$0.118}} \\
\bottomrule
\end{tabular}
}
\end{table*}

\begin{table}[t]
\centering
\caption{\textbf{Efficiency comparison} on the short-video benchmark (241-frame clips). All measurements use one NVIDIA H100 GPU. Wall-clock FPS counts generated pixel frames per second. SCAIL-2 uses its accelerated 8-step LoRA configuration. Baseline times cover an entire clip, whereas UniSwap's values marked by $\dagger$ are per block (3 latent frames or 24 pixel frames); per-step times are therefore not directly comparable across the two settings.}
\label{tab:efficiency}
\vspace{2mm}
\resizebox{\linewidth}{!}{
\begin{tabular}{l|cc|c|c}
\toprule
Method & Steps & Infer.\ Time (s) & Time/Step (s) & FPS $\uparrow$ \\
\midrule
VACE~\cite{vace} & 50 & 482.56 & 9.65 & 0.499 \\
Wan-Animate~\cite{wananimate} & 20 & 176.34 & 8.82 & 1.367 \\
HunyuanCustom~\cite{hunyuancustom} & 50 & 703.34 & 14.07 & 0.343 \\
SCAIL-2~\cite{scail2} & 8 & 190.98 & 23.87 & 1.262 \\
MoCha~\cite{mocha} & 30 & 1800.94 & 60.03 & 0.134 \\
\midrule
\textbf{UniSwap} & 3$^\dagger$ & \textbf{1.76}$^\dagger$ & 0.59$^\dagger$ & \textbf{13.6} \\
\bottomrule
\end{tabular}
}
\end{table}

\begin{table*}[!ht]
\centering
\caption{\textbf{Ablation on training stages and the condition positional encoding offset} (short-video benchmark). The best result in each column is in \textbf{bold} and the second best is \underline{underlined}.}
\label{tab:ablation_stages}
\vspace{2mm}
\resizebox{\textwidth}{!}{
\begin{tabular}{l|cc|ccc|ccccc}
\toprule
\multirow{2}{*}{Setting} & \multicolumn{2}{c|}{A--V Sync} & \multicolumn{3}{c|}{Video Quality} & \multicolumn{5}{c}{Voice Quality} \\
\cmidrule(lr){2-3} \cmidrule(lr){4-6} \cmidrule(lr){7-11}
& Sync-C $\uparrow$ & Sync-D $\downarrow$ & ASE $\uparrow$ & IQA $\uparrow$ & DINO-S $\uparrow$ & SIG $\uparrow$ & BAK $\uparrow$ & OVRL $\uparrow$ & SECS $\uparrow$ & SSIM $\uparrow$ \\
\midrule
Stage 1 (In-context) & \textbf{5.272{\scriptsize$\pm$1.510}} & \textbf{9.107{\scriptsize$\pm$0.941}} & \textbf{2.253{\scriptsize$\pm$0.289}} & \textbf{3.922{\scriptsize$\pm$0.325}} & \textbf{0.635{\scriptsize$\pm$0.132}} & \underline{3.476{\scriptsize$\pm$0.422}} & \textbf{3.619{\scriptsize$\pm$0.665}} & \textbf{3.029{\scriptsize$\pm$0.498}} & \textbf{0.782{\scriptsize$\pm$0.059}} & 0.213{\scriptsize$\pm$0.151} \\
Stage 2 (Teacher forcing) & \underline{4.620{\scriptsize$\pm$1.187}} & \underline{9.581{\scriptsize$\pm$0.752}} & \underline{2.233{\scriptsize$\pm$0.304}} & \underline{3.893{\scriptsize$\pm$0.332}} & 0.623{\scriptsize$\pm$0.134} & 3.349{\scriptsize$\pm$0.578} & 3.059{\scriptsize$\pm$0.736} & 2.687{\scriptsize$\pm$0.549} & 0.681{\scriptsize$\pm$0.071} & \textbf{0.480{\scriptsize$\pm$0.156}} \\
Stage 3 (Self-forcing DMD) & 3.633{\scriptsize$\pm$1.236} & 10.304{\scriptsize$\pm$0.849} & 2.097{\scriptsize$\pm$0.238} & 3.758{\scriptsize$\pm$0.318} & \underline{0.629{\scriptsize$\pm$0.136}} & \textbf{3.486{\scriptsize$\pm$0.327}} & \underline{3.563{\scriptsize$\pm$0.543}} & \underline{2.988{\scriptsize$\pm$0.366}} & \underline{0.730{\scriptsize$\pm$0.064}} & \underline{0.269{\scriptsize$\pm$0.161}} \\
\midrule
Stage 2 w/o condition PE offset & 1.738{\scriptsize$\pm$1.737} & 11.843{\scriptsize$\pm$1.606} & 2.152{\scriptsize$\pm$0.352} & 3.472{\scriptsize$\pm$0.600} & 0.463{\scriptsize$\pm$0.166} & 3.238{\scriptsize$\pm$0.728} & 3.280{\scriptsize$\pm$0.924} & 2.767{\scriptsize$\pm$0.702} & 0.624{\scriptsize$\pm$0.086} & 0.103{\scriptsize$\pm$0.142} \\
\bottomrule
\end{tabular}
}
\end{table*}

\begin{table*}[t]
\centering
\caption{\textbf{Ablation on Feature-RoPE Decomposition} (long-video benchmark, per-segment metrics). The best result in each segment is in \textbf{bold} and the second best is \underline{underlined}.}
\label{tab:ablation_rerope}
\vspace{2mm}
\resizebox{\textwidth}{!}{
\begin{tabular}{l|ccc|ccc|ccc}
\toprule
\multirow{2}{*}{Setting} & \multicolumn{3}{c|}{0--20\,s} & \multicolumn{3}{c|}{20--40\,s} & \multicolumn{3}{c}{40--60\,s} \\
\cmidrule(lr){2-4} \cmidrule(lr){5-7} \cmidrule(lr){8-10}
& ASE $\uparrow$ & IQA $\uparrow$ & DINO-S $\uparrow$ & ASE $\uparrow$ & IQA $\uparrow$ & DINO-S $\uparrow$ & ASE $\uparrow$ & IQA $\uparrow$ & DINO-S $\uparrow$ \\
\midrule
\textbf{UniSwap (full)} & \textbf{2.224{\scriptsize$\pm$0.226}} & \textbf{3.966{\scriptsize$\pm$0.331}} & \underline{0.596{\scriptsize$\pm$0.122}} & \textbf{2.236{\scriptsize$\pm$0.189}} & \textbf{4.001{\scriptsize$\pm$0.276}} & \textbf{0.590{\scriptsize$\pm$0.126}} & \textbf{2.259{\scriptsize$\pm$0.191}} & \textbf{4.032{\scriptsize$\pm$0.254}} & \textbf{0.596{\scriptsize$\pm$0.118}} \\
\midrule
\quad w/o Window-Bounded RoPE & \underline{2.115{\scriptsize$\pm$0.226}} & \underline{3.763{\scriptsize$\pm$0.228}} & \textbf{0.599{\scriptsize$\pm$0.120}} & 2.083{\scriptsize$\pm$0.227} & \underline{3.500{\scriptsize$\pm$0.153}} & \underline{0.546{\scriptsize$\pm$0.093}} & 2.083{\scriptsize$\pm$0.213} & \underline{3.390{\scriptsize$\pm$0.141}} & \underline{0.517{\scriptsize$\pm$0.071}} \\
\quad w/o Reference Re-anchoring & 2.106{\scriptsize$\pm$0.197} & 3.741{\scriptsize$\pm$0.314} & 0.595{\scriptsize$\pm$0.114} & \underline{2.111{\scriptsize$\pm$0.225}} & 3.399{\scriptsize$\pm$0.207} & 0.522{\scriptsize$\pm$0.094} & \underline{2.140{\scriptsize$\pm$0.227}} & 3.208{\scriptsize$\pm$0.246} & 0.491{\scriptsize$\pm$0.084} \\
\quad w/o Adaptive Sink Block & 2.083{\scriptsize$\pm$0.173} & 3.734{\scriptsize$\pm$0.313} & 0.589{\scriptsize$\pm$0.115} & 2.007{\scriptsize$\pm$0.143} & 3.336{\scriptsize$\pm$0.164} & 0.530{\scriptsize$\pm$0.087} & 2.099{\scriptsize$\pm$0.203} & 3.174{\scriptsize$\pm$0.182} & 0.499{\scriptsize$\pm$0.070} \\
\bottomrule
\end{tabular}
}
\end{table*}

\subsection{Experimental Setup}

\noindent\textbf{Training Data.}
We train UniSwap on AVSpeech~\cite{avspeech}, a large-scale corpus of talking videos with clean speech from diverse speakers. Paired training tuples are synthesized from these unpaired clips using the swap-and-reconstruct pipeline described in Sec.~\ref{sec:data}. Both training and inference support three aspect-ratio buckets: $512{\times}512$, $416{\times}704$, and $704{\times}416$ pixels. Training clips contain 241 frames (approximately 9.6 seconds at 25 fps).

\noindent\textbf{Implementation Details.}
All three stages share the frozen LTX-2.3 base model and train Low-Rank Adaptation (LoRA) adapters~\cite{lora} in bf16 mixed precision on 8 GPUs with FSDP. Stage 1 applies LoRA of rank 128 to the attention projections and is optimized with AdamW (learning rate $1{\times}10^{-4}$, linear schedule) for 50{,}000 steps. Stage 2 applies LoRA of rank 128 to the attention projections and feed-forward layers of both modalities as well as the cross-modal attention, and is optimized with AdamW (learning rate $1{\times}10^{-4}$) for 50{,}000 steps. Stage 3 uses LoRA of rank 128 for all three roles and is optimized with AdamW (both generator and critic learning rates $1{\times}10^{-5}$, $\beta_1{=}0$, $\beta_2{=}0.999$) for 20{,}000 steps, with the critic updated five times per generator update. Inference runs on a single NVIDIA H100 GPU with the cache configuration described in Sec.~\ref{sec:rerope}. All methods are evaluated on the same source clips, reference identities, resolutions, and frame rates, and we report the mean $\pm$ standard deviation over the evaluation clips.

\noindent\textbf{Evaluation Benchmarks.}
We evaluate on two benchmarks. (1) A short-video benchmark of 100 clips with an approximate duration of 10 seconds, collected from AVSpeech speakers disjoint from the training set and categorized by body visibility into head, half-body, and full-body shots according to the detected pose; reference portraits are randomly collected photorealistic human images with diverse poses, including head, half-body, and full-body shots with frontal and profile faces. (2) A long-video benchmark of 20 web-crawled talking videos of 1 minute in duration, with the same diversity of body shots and reference images, on which metrics are computed independently for each 20-second segment to expose temporal drift.

\noindent\textbf{Baselines.}
We compare against three categories of methods. (1) \textit{Video replacement}: MoCha~\cite{mocha}, Wan-Animate~\cite{wananimate}, VACE~\cite{vace}, HunyuanCustom~\cite{hunyuancustom}, and SCAIL-2~\cite{scail2}, which replace the visual identity but do not replace the speaker identity. We therefore pair every visual output with speech converted by the same Seed-VC backend and evaluate the resulting cascade as the direct competitor to joint replacement. We select Seed-VC as the strongest open-source voice-conversion method in our evaluation: it achieves the best SIG, BAK, OVRL, and SECS scores among the evaluated converters (Table~\ref{tab:main_results}) and is also used to synthesize the source audio in our training pipeline. Using a common backend across all video-only methods further isolates differences in their visual components. (2) \textit{Voice conversion}: Seed-VC~\cite{seedvc}, CosyVoice~\cite{cosyvoice}, and OpenVoice~\cite{openvoice}, which convert the voice while keeping the original video. (3) \textit{Ours}: UniSwap, the final streaming generator; the individual training stages are compared separately in the ablation study (Sec.~\ref{sec:ablation}).

\noindent\textbf{Metrics.}
We evaluate three aspects. (1) \textit{Audio-visual synchronization}: Sync-C (confidence, higher is better) and Sync-D (feature distance, lower is better) from SyncNet~\cite{syncnet}. (2) \textit{Video quality}: aesthetic (ASE) and image-quality (IQA) scores computed by Q-Align~\cite{qalign}, and DINO-S~\cite{dino}, the cosine similarity between features of the generated frames and reference image for identity preservation. (3) \textit{Voice quality} (following Seed-VC~\cite{seedvc}): SIG, BAK, and OVRL from DNSMOS~\cite{dnsmos} for perceptual speech quality, the speaker encoder cosine similarity (SECS) to the reference voice~\cite{resemblyzer}, and the structural similarity (SSIM)~\cite{ssim} between the mel spectrograms of the generated and source audio, which measures how well the converted speech preserves the original content.

\subsection{Qualitative Results}

Fig.~\ref{fig:qual_short} shows representative examples from the short-video benchmark. Video replacement baselines transfer the reference appearance but require a separate voice-conversion stage; voice conversion baselines modify only the waveform and cannot change the appearance. UniSwap jointly replaces both modalities while preserving the source composition and motion: the generated character follows the source pose and background, matches the reference appearance, and maintains lip motion synchronized with the generated speech while matching the reference vocal timbre.

Fig.~\ref{fig:qual_long} further illustrates long-duration generation. SCAIL-2 and Wan-Animate exhibit increasing identity drift and visual artifacts in later segments, whereas UniSwap remains stable over the full minute, consistent with the per-segment metrics in Table~\ref{tab:long_video}.

\subsection{Quantitative Results}

Table~\ref{tab:main_results} presents the short-video comparison. UniSwap obtains the highest Sync-C (3.633) and lowest Sync-D (10.304) among the evaluated replacement pipelines. Unlike cascades that generate lip motion without access to the converted speech, UniSwap generates both modalities jointly. Its visual identity similarity matches the strongest baseline within 0.001 DINO-S (0.629 versus 0.630), although its aesthetic and image-quality scores remain below MoCha and SCAIL-2. For speech, UniSwap achieves a SIG score of 3.486, close to Seed-VC (3.489), but remains below the best voice-conversion result on BAK, OVRL, SECS, and SSIM. These results reflect the trade-off between unified joint replacement and separately optimized single-modality systems.

Table~\ref{tab:long_video} compares one-minute generations by segment. UniSwap's IQA remains between 3.966 and 4.032, and its DINO-S remains between 0.590 and 0.596. It achieves the highest identity similarity in all three segments, while SCAIL-2 retains higher aesthetic and image-quality scores. Wan-Animate's IQA decreases from 3.766 to 3.628; SCAIL-2's DINO-S decreases from 0.566 to 0.517. The results indicate that UniSwap preserves identity more consistently over the evaluated duration, but they do not establish superiority on every visual-quality dimension.

Table~\ref{tab:efficiency} compares generation efficiency. UniSwap generates a 24-frame block in 1.76 seconds, corresponding to 13.6 wall-clock FPS. This is approximately 10$\times$ faster than the fastest evaluated baseline (Wan-Animate at 1.367 FPS) and about 100$\times$ faster than MoCha. The reported SCAIL-2 result uses its accelerated 8-step LoRA rather than the 40-step default sampler. The gain comes from processing only the current block against cached context rather than updating the full sequence at every denoising step. Because 13.6 FPS remains below the 25-FPS playback rate used in our experiments, the current implementation supports streaming generation but not real-time playback; further systems optimization is required. Per-step time is also not directly comparable across rows because a baseline step updates an entire clip, whereas a UniSwap step updates one block.

\subsection{Ablation Studies}
\label{sec:ablation}

\noindent\textbf{Ablation on Training Stages.}
Table~\ref{tab:ablation_stages} compares the three training stages. Stage 1 obtains the strongest synchronization, video-quality, and speaker-similarity scores. Stage 2 retains comparable visual quality while enabling block-causal generation. Stage 3 reduces denoising from 30 to 3 steps per block; compared with Stage 2, it improves DINO-S, SIG, BAK, OVRL, and SECS, but decreases synchronization, ASE, IQA, and SSIM.

\noindent\textbf{Ablation on PE Offset.}
Removing the condition positional encoding offset from Stage 2 reduces Sync-C from 4.620 to 1.738 and DINO-S from 0.623 to 0.463, supporting its role in audio-video synchronization and reference-identity preservation.

\noindent\textbf{Ablation on Feature-RoPE Decomposition.}
Table~\ref{tab:ablation_rerope} ablates the three positional components on the long-video benchmark. Removing Window-Bounded RoPE, Reference Re-anchoring, or the Adaptive Sink Block leads to progressive degradation across segments, with the effect compounding over time: without Reference Re-anchoring, for example, IQA drops from 3.741 to 3.208 and DINO-S from 0.595 to 0.491 by the final segment.

Additional qualitative ablations are provided in supplementary Fig.~\ref{fig:ablation_qual}.

\section{Conclusion}
\label{sec:conclusion}

We presented UniSwap, the first framework for streaming joint audio-visual identity replacement in talking videos. UniSwap formulates appearance and voice transfer as a unified conditional generation task and constructs aligned supervision through swap-and-reconstruct data synthesis. In-context Pretraining learns joint visual and vocal identity replacement, Conditional Streaming Adaptation introduces the Decoupled Streaming Conditioning Mask for block-causal generation, and Efficient Self-forcing DMD reduces sampling from 30 to 3 denoising steps per block. Efficient Multi-LoRA Switching enables memory-efficient distillation on a shared backbone, while Feature-RoPE Decomposition supports bounded-cache inference for long-form generation. Experiments show that UniSwap achieves stronger audio-visual synchronization than the evaluated cascaded systems, remains competitive in identity preservation and perceptual quality, and maintains stable visual identity over one-minute sequences.

\clearpage
{
    \small
    \bibliographystyle{ieeenat_fullname}
    \bibliography{main}
}

\clearpage
\hypersetup{pageanchor=false}
\setcounter{page}{1}
\maketitlesupplementary

\section{Qualitative Ablation}

\begin{figure*}[!ht]
    \centering
    \includegraphics[width=\textwidth]{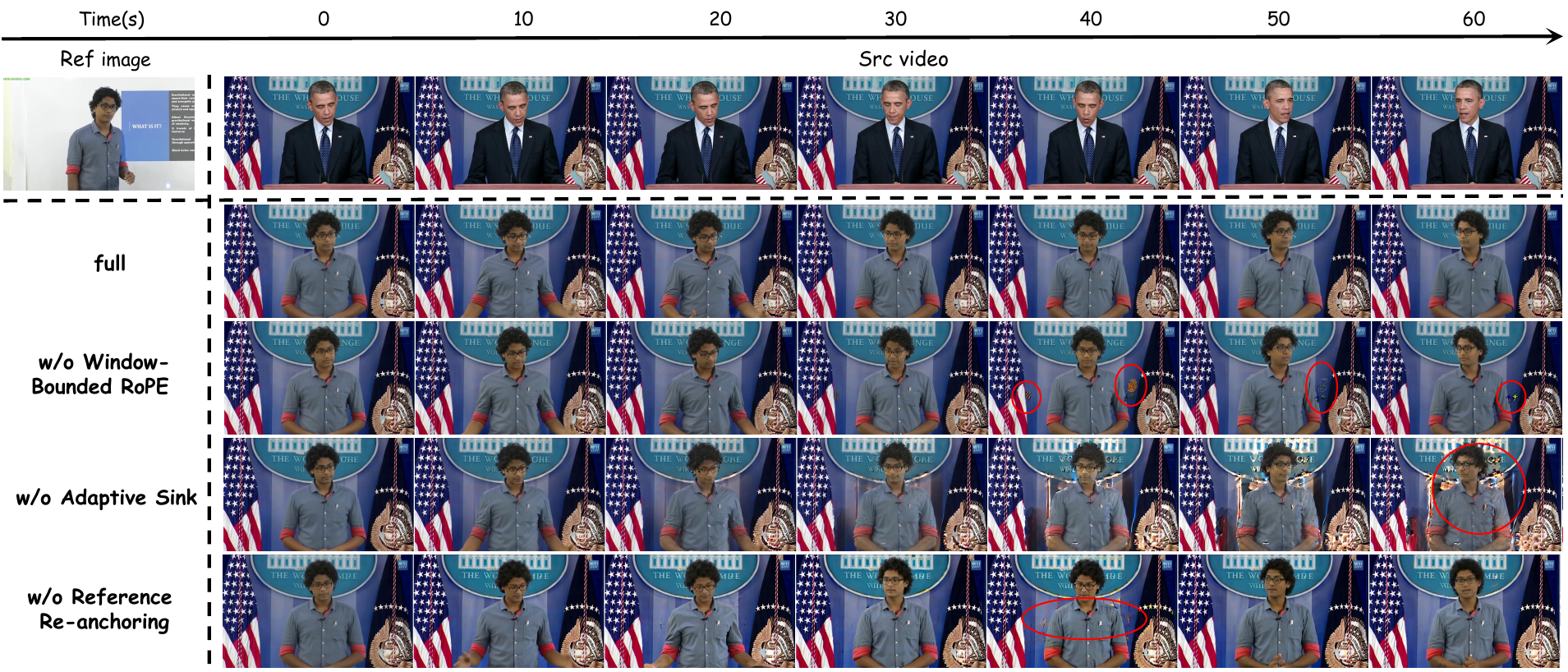}
    \caption{\textbf{Qualitative ablation on Feature-RoPE Decomposition.} Frames are sampled every 10 seconds from one-minute generations. Removing any component leads to visible identity drift and artifacts in later segments, whereas the full model remains more consistent throughout the sequence. Zoom in for details.}
    \label{fig:ablation_qual}
\end{figure*}

Figure~\ref{fig:ablation_qual} complements the quantitative ablation in the main paper by visualizing the effect of each component of Feature-RoPE Decomposition over one-minute generations. The ablated variants exhibit increasing identity drift and visual artifacts in later segments, whereas the full model remains more consistent. Together with Table~\ref{tab:ablation_rerope}, these results support the roles of bounded coordinates, reference re-anchoring, and the adaptive sink block in reducing long-horizon drift.

\section{KV-Cached Streaming Inference}
\label{sec:supp_kv_cache}

Algorithm~\ref{alg:kv_cache} details the blockwise inference procedure used by Stages 2 and 3. The reference cache persists throughout generation, source keys and values are temporary, and completed target blocks are committed to the clean-history cache.

\begin{algorithm}[H]
\caption{KV-Cached Streaming Inference}
\label{alg:kv_cache}
\begin{algorithmic}[1]
\REQUIRE Reference tokens $R$; source blocks $\{S_i\}_{i=0}^{N-1}$; reference region $\mathcal{R}$; history $\mathcal{H}$ (sink + rolling)
\STATE \textbf{Prefill reference:} forward $R$ once and write its key/value tensors into $\mathcal{R}$ \COMMENT{never evicted}
\FOR{$i = 0$ \TO $N-1$}
    \STATE \textbf{Prefill source:} forward $S_i$ at the current slot and temporarily cache its key/value tensors
    \STATE \textbf{Denoise:} iteratively denoise $B_i$ in read-only mode, attending to $\mathcal{R}$, $\mathcal{H}$, and $S_i$
    \STATE \textbf{Commit:} forward the denoised $B_i$ as clean context, and append its unrotated keys and values to the sink region if $i = 0$, or to the rolling region otherwise
    \STATE \textbf{Shift:} once the rolling region is full, evict its oldest block, shift the retained blocks, and reapply RoPE to their cached keys at bounded local coordinates
    \STATE \textbf{Rollback:} discard the temporary key/value tensors of $S_i$
\ENDFOR
\RETURN Generated blocks $\{B_i\}_{i=0}^{N-1}$
\end{algorithmic}
\end{algorithm}

\section{User Study}
\label{sec:user_study}

We conducted a blinded user study with 30 participants. The study compared UniSwap with four video-replacement baselines, each paired with Seed-VC following the cascade protocol in Table~\ref{tab:main_results}. Each participant evaluated anonymized outputs from all five methods on four source clips and rated appearance identity, voice identity, lip synchronization, and naturalness on a five-point Likert scale. Method order was randomized independently for each participant.

\begin{table}[t]
\centering
\caption{\textbf{User-study results.} Ratings are mean scores on a five-point Likert scale.}
\label{tab:user_study}
\vspace{2mm}
\resizebox{\linewidth}{!}{
\begin{tabular}{l|cccc}
\toprule
Method & Appearance ID $\uparrow$ & Voice ID $\uparrow$ & Lip Sync $\uparrow$ & Naturalness $\uparrow$ \\
\midrule
Wan-Animate + Seed-VC & 3.85 & 4.04 & 3.42 & 3.77 \\
SCAIL-2 + Seed-VC & 4.03 & 4.05 & 3.67 & 3.85 \\
MoCha + Seed-VC & 3.91 & \textbf{4.08} & 3.54 & 3.89 \\
HunyuanCustom + Seed-VC & 3.64 & 3.98 & 3.28 & 3.61 \\
\textbf{UniSwap} & \textbf{4.16} & 3.87 & \textbf{4.11} & \textbf{3.96} \\
\bottomrule
\end{tabular}
}
\end{table}

UniSwap receives the highest ratings for appearance identity, lip synchronization, and naturalness.

\section{Additional Qualitative Results}
\label{sec:supp_qualitative}

We provide additional qualitative results for both short and long source videos. In all figures, each example contains a reference image and reference voice clip, a source video and its audio, and the joint audio-video output produced by UniSwap. The red and blue waveforms denote the source and generated audio, respectively. These examples complement the quantitative evaluation in the main paper by covering diverse identities, poses, gestures, clothing, backgrounds, and recording conditions.

\subsection{Short-Video Results}
\label{sec:supp_short}

\begin{figure*}[t]
    \centering
    \includegraphics[width=0.94\textwidth]{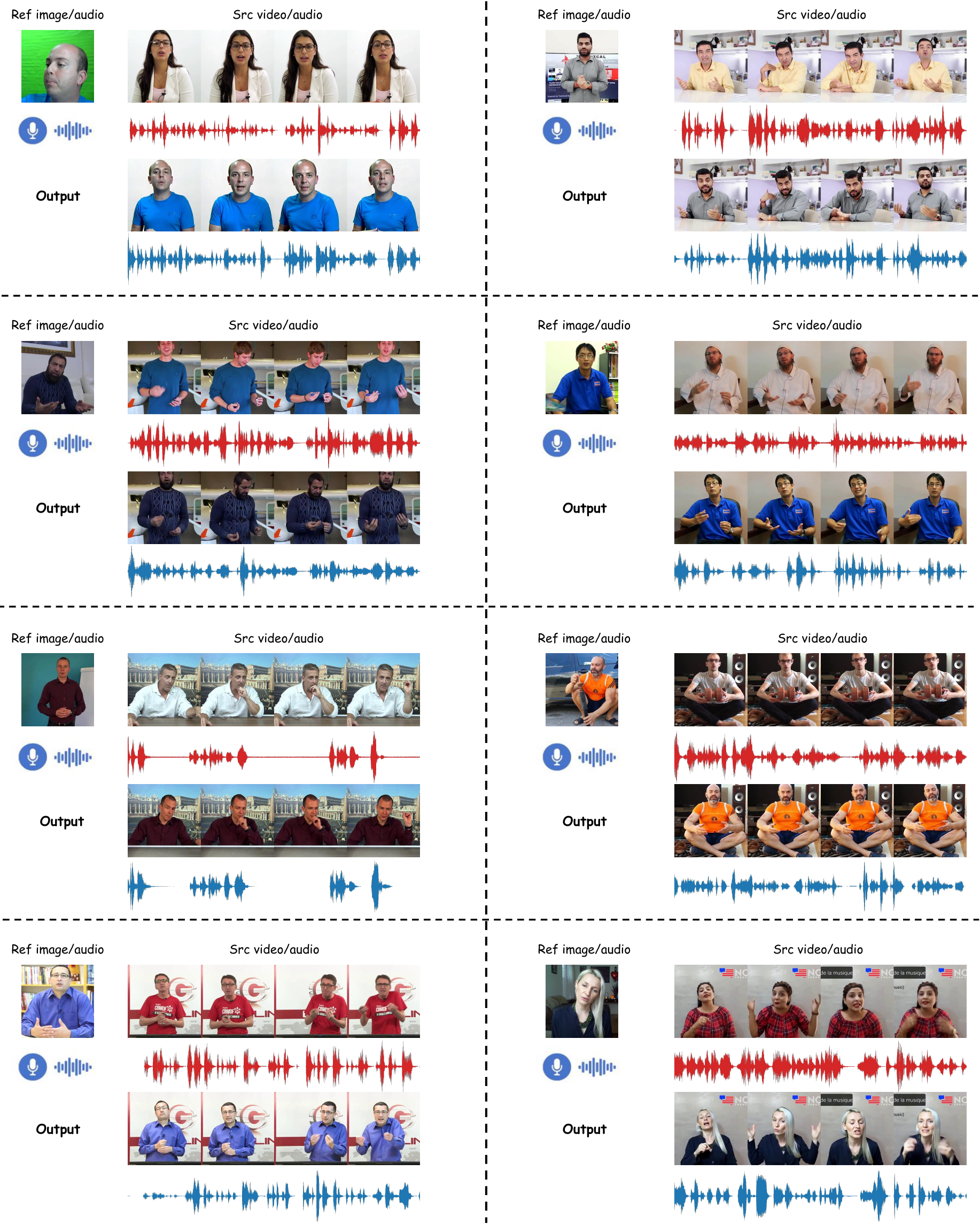}
    \caption{\textbf{Additional qualitative results on short videos (Part I).} For each example, the left column provides the reference image and reference voice, while the right column shows sampled frames and the waveform of the source video/audio followed by the UniSwap output. Red waveforms indicate source audio and blue waveforms indicate generated audio. UniSwap changes the visual and vocal identity according to the references while preserving the source scene composition and motion.}
    \label{fig:supp_short_1}
\end{figure*}

\begin{figure*}[t]
    \centering
    \includegraphics[width=0.98\textwidth]{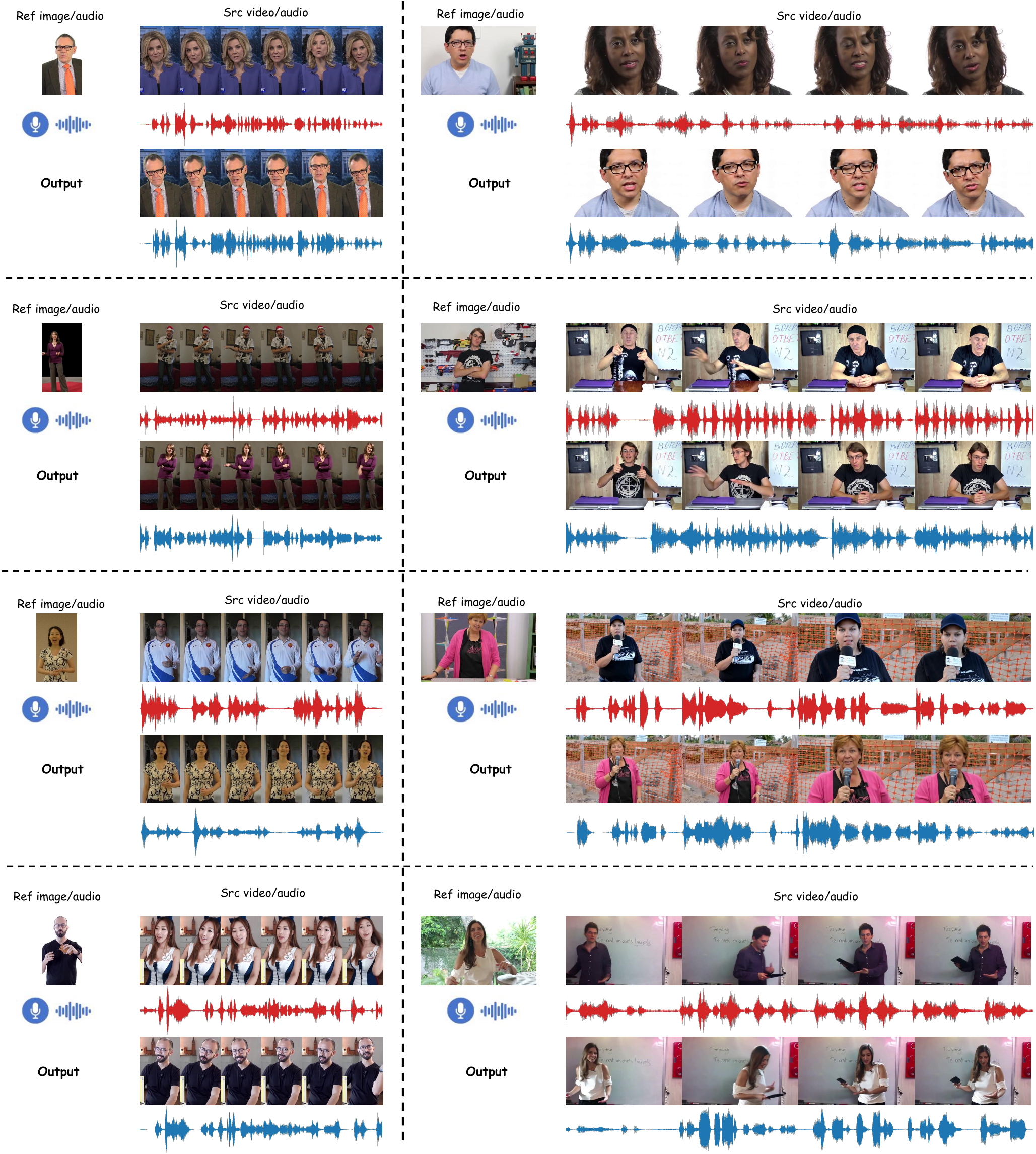}
    \caption{\textbf{Additional qualitative results on short videos (Part II).} The examples follow the layout of Fig.~\ref{fig:supp_short_1} and cover additional identities, viewpoints, gestures, and backgrounds. Red and blue denote the source- and generated-audio waveforms, respectively. The generated frames adopt the reference appearance while following the pose, expression, framing, and scene content of the source video.}
    \label{fig:supp_short_2}
\end{figure*}

Figures~\ref{fig:supp_short_1} and~\ref{fig:supp_short_2} present 16 additional short-video examples. Across these cases, UniSwap transfers the appearance specified by the reference image while retaining the source composition, body motion, and facial activity. The examples include cross-gender replacement, varied camera framing, substantial hand and upper-body motion, and both simple and cluttered backgrounds. The corresponding generated-audio waveforms are shown alongside the visual outputs to illustrate that the two modalities are produced jointly for the full source sequence.

\subsection{Long-Video Results}
\label{sec:supp_long}

\begin{figure*}[t]
    \centering
    \includegraphics[width=\textwidth]{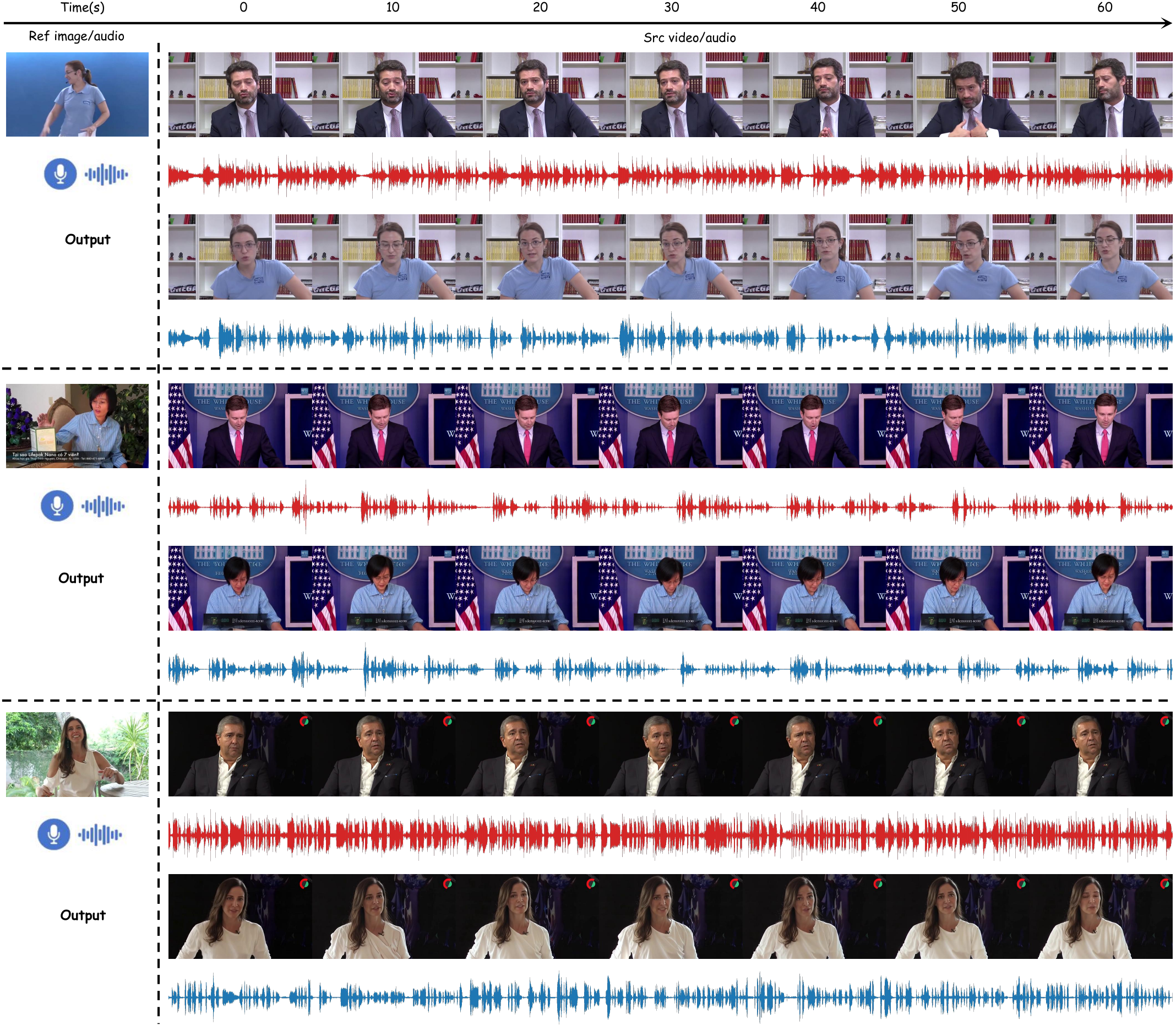}
    \caption{\textbf{Additional qualitative results on one-minute videos.} Frames are sampled every 10 seconds from 0 to 60 seconds. Each row group shows the reference image and voice, the source video/audio (red waveform), and the UniSwap output (blue waveform). Across all three sequences, the generated character remains consistent with the reference identity throughout the minute while retaining the source scene and temporal performance.}
    \label{fig:supp_long}
\end{figure*}

Figure~\ref{fig:supp_long} shows three additional one-minute generations sampled at 10-second intervals. The output identity remains visually consistent from the beginning to the end of each sequence despite continuous autoregressive generation. At the same time, the outputs preserve the source background, camera framing, and time-varying facial and upper-body motion. These results provide further qualitative evidence that the bounded cache coordinates and persistent identity context used by UniSwap mitigate long-horizon identity drift.

\section{Limitations}

UniSwap currently targets single-speaker talking videos; multi-speaker scenes, occlusions, and complex interactions remain challenging. Facial expressions are driven automatically by the audio condition rather than controlled explicitly, so the current model does not support independent expression editing or arbitrary user-specified expression control.

\section{Broader Impact}

Audio-video character replacement can support filmmaking, localization, and accessibility, but it also increases the risk of impersonation, non-consensual media, and misinformation. Deployment should require consent and provenance mechanisms, visible disclosure where appropriate, access controls, and compatibility with forensic detection tools.

\end{document}